\pdfoutput=1

\pdfoutput=1

\documentclass[11pt]{article}

\usepackage[]{acl}

\usepackage{times}
\usepackage{latexsym}
\usepackage{hyperref}
\usepackage{url}
\usepackage{multirow}
\usepackage{graphicx}
\usepackage{tabularx}
\usepackage{adjustbox}
\usepackage{times}
\usepackage{latexsym}
\usepackage{booktabs}
\usepackage{tabto}
\usepackage{float}
\usepackage{wrapfig}
\usepackage{xspace}
\usepackage{enumitem}

\usepackage[T1]{fontenc}

\usepackage[utf8]{inputenc}

\usepackage{microtype}

\usepackage{inconsolata}

\newcommand{\taskname}{\texttt{$X$-Shot}\xspace}
\newcommand{\modelname}{\texttt{BinBin}\xspace}
\newcommand{\instrudata}{Super-NaturalInstruction\xspace}

\title{\taskname: A Unified System to Handle \texttt{Frequent}, \texttt{Few-shot} and \texttt{Zero-shot} Learning Simultaneously in Classification}

\author{Hanzi Xu$^1$ \ \ \ Muhao Chen$^2$ \ \ \ Lifu Huang$^3$ \ \ \ Slobodan Vucetic$^1$ \ \ \ Wenpeng Yin$^4$\\
$^1$Temple University \ \ \ $^2$University of California, Davis \ \ \ $^3$Virginia Tech \ \ \ $^4$Penn State University\\
 \texttt{\{hanzi.xu, slobodan.vucetic\}@temple.edu} \ \ \ 
\texttt{wenpeng@psu.edu}
}

\begin{document}
\maketitle
\begin{abstract}
In recent years, few-shot and zero-shot learning, which learn to predict labels with limited annotated instances, have garnered significant attention. Traditional approaches often treat frequent-shot (freq-shot; labels with abundant instances), few-shot, and zero-shot learning as distinct challenges, optimizing systems for just one of these scenarios. Yet, in real-world settings, label occurrences vary greatly. Some of them might appear thousands of times, while others might only appear sporadically or not at all. For practical deployment, it is crucial that a system can adapt to any label occurrence. We introduce a novel classification challenge: \taskname, reflecting a real-world context where freq-shot, few-shot, and zero-shot labels co-occur without predefined limits. Here, $X$ can span from 0 to +$\infty$. The crux of \taskname centers on open-domain generalization and devising a system versatile enough to manage various label scenarios. To solve \taskname, we propose \modelname~(\textbf{b}inary \textbf{in}ference \textbf{b}ased on \textbf{in}struction following) that leverages the \emph{Indirect Supervision} from a large collection of NLP tasks via instruction following, bolstered by \emph{Weak Supervision} provided by large language models. \modelname~surpasses previous state-of-the-art techniques on three benchmark datasets across multiple domains. To our knowledge, this is the first work addressing \taskname learning, where $X$ remains variable.\footnote{Code and data are publicly available at \url{https://github.com/xhz0809/X-shot}.}

\end{abstract}

\section{Introduction}

For classification problems, the distribution of label occurrences in real-world scenarios often varies widely, with some labels appearing frequently (frequent-shot), others infrequently (few-shot), and some not at all (zero-shot). Given this variability, it becomes imperative to craft learning systems adept at managing labels across the full frequency spectrum. Regrettably, current few-shot systems often fall short when confronted with zero-shot challenges \citep{DBLP:conf/emnlp/ZhangZHY22, DBLP:conf/acl/CuiHDHL22, DBLP:conf/icml/ZhaoWFK021}. In contrast, zero-shot systems, while adept in their domain, cannot fully benefit from the potential advantages of annotations when available \citep{DBLP:conf/naacl/ZhangLG19, obamuyide2018zero, DBLP:conf/emnlp/YinHR19,DBLP:journals/corr/abs-2210-14299}. Thus, developing the skill to manage all possible label occurrences simultaneously is crucial for systems that are intended for practical use.

In this work, we introduce a more challenging and practically useful task:\taskname learning. This task mirrors real-world environments where label occurrence spans a continuum, seamlessly incorporating frequent-shot, few-shot, and zero-shot instances, all without a priori constraints. In this paradigm, variable $X$, the number of times each label is seen during the training, is unbounded, ranging freely within the interval [0, +$\infty$). At the heart of \taskname lies the objective of attaining open-domain generalization and architecting a system resilient across a plethora of label scenarios.

Tackling \taskname spawns two core technical conundrums: ($\mathcal{Q}_1$) How can one identify suitable sources of \emph{Indirect Supervision} \citep{DBLP:conf/acm/0001CZNCR23} in few-shot and zero-shot settings, given the notable scarcity of annotations. ($\mathcal{Q}_2$) Traditional multi-class classifiers struggle with the diversity in label sizes across tasks, frequently requiring customized classification heads for each variation. Here, the challenge is formulating a cohesive system capable of effectively adapting to labels of diverse sizes. 


To address $\mathcal{Q}_1$, we identify the most effective source of \emph{Indirect Supervision} as being from instruction tuning datasets, such as \instrudata \citep{DBLP:conf/emnlp/WangMAKMNADASPK22}. These datasets primarily contain various NLP tasks enriched with textual instructions. Our method trains the model on these datasets, aiming for robust generalization to the unseen \taskname~task when supplemented with pertinent instructions, especially for the low-shot (few-shot and zero-shot) labels. For $\mathcal{Q}_2$, we advocate a triplet-oriented binary classifier. This classifier functions by accepting a triplet of ($\texttt{instruction}$, $\texttt{input}$, $\texttt{label}$), anticipating a binary response (``Yes'' or ``No'') that confirms the suitability of the $\texttt{label}$ for the specified $\texttt{input}$ under the given \texttt{instruction}. Such a triplet-oriented classifier acts as a cohesive architecture that manages text classification tasks with labels of varied sizes. By combining solutions for both $\mathcal{Q}_1$ and $\mathcal{Q}_2$, we forge a holistic framework, \modelname~(\textbf{b}inary \textbf{in}ference \textbf{b}ased on \textbf{in}struction following). 

There are, however, no existing datasets that explicitly cater to this challenge. To evaluate our system, we turn to three representative classification tasks: relation classification, event detection, and argument role identification. We recompile their associated datasets: \textit{FewRel} \citep{DBLP:conf/emnlp/HanZYWYLS18}, \textit{MAVEN} \citep{DBLP:conf/emnlp/WangWHJHLLLLZ20}, and \textit{RAMS} \citep{DBLP:conf/acl/EbnerXCRD20} and simultaneously include frequent-shot, few-shot, and zero-shot instances. Sourced from diverse domains (Wikipedia, news articles, etc.), and featuring vast label counts (ranging from 30 to 78), these datasets pose a formidable challenge to contemporary text classification systems. Moreover, the \textit{MAVEN} dataset uniquely integrates a ``None'' label, further amplifying the realistic nature of the task. Experiments on multiple model scales and architectures reveal our system's resilience across datasets, consistently outperforming leading baselines, including GPT-3.5.

Our contributions can be summarized as follows: (i) We introduce \taskname, a hitherto under-explored, open-domain open-shot text classification problem that mirrors real-world complexities. (ii) We innovate a unique problem setting that reframes any text classification challenge into a binary classification task, adaptable to any number of label sizes and occurrences. (iii) Our \modelname, harnessing the potential of instruction-following datasets, excels past existing approaches, demonstrating versatility across various domains, label magnitudes, and classification paradigms.

\section{Related Work}

\paragraph{Data Imbalance}  The topic of data-imbalanced NLP Tasks is first discussed in the context of binary classification datasets, where the negative-to-positive ratio ranges from 5 to 200 \citep{DBLP:conf/acl/LiSMLWL20}. Subsequent works have extended this to multi-class classification settings with a long-tail distribution, where a subset of labels occurs in less than 5\% of the training data \cite{DBLP:conf/nips/CaoWGAM19, DBLP:conf/aaai/XuXLLJY23}. Two common solutions to this problem are reweighting the loss function and resampling the data in mini-batches \cite{DBLP:conf/acl/LiSMLWL20,DBLP:conf/nips/CaoWGAM19, DBLP:conf/aaai/XuXLLJY23,DBLP:journals/nn/BudaMM18,DBLP:conf/iri/PouyanfarCS18}. Even though the data imbalance/long-tail problem also tackles different label occurrences, this setting differs from the \taskname problem in three dimensions: i) the presence of zero-shot labels in our setting; ii) the inclusion of a “None” class in the test set, representing cases where none of the labels fit; iii) prior work addressed different imbalance/long-tail problems with separate systems (a system for task/domain A could not be applied to another task/domain), whereas we are modeling these problems within a unified system. 

\paragraph{Indirect Supervision}
There has been a burgeoning interest in \emph{Indirect Supervision} \citep{DBLP:conf/acm/0001CZNCR23} in recent years. Here, easily available signals from relevant tasks (source tasks) are used to aid in learning the target task. Using the entailment task for \emph{Indirect Supervision} in zero-shot classification was first proposed by \citep{DBLP:conf/emnlp/YinHR19} and has since been adapted for a variety of NLP tasks, including few-shot intent identification \citep{DBLP:conf/emnlp/ZhangHLWWYSX20,xu-etal-2023-dense}, event argument extraction \citep{DBLP:conf/naacl/SainzGLMA22} and relation extraction \citep{DBLP:conf/emnlp/LuHZMC22}. Beyond entailment, knowledge from areas like question answering \citep{DBLP:conf/acl/YinRX21}, summarization \citep{DBLP:conf/emnlp/LuHZMC22} and dense retrievers \citep{xu-etal-2023-dense} has been incorporated. However, previous \emph{Indirect Supervision} is collected from a single source task. In contrast, our work is inspired by recent studies in instruction learning observing the efficacy of NLP models when given task instructions and their ability to generalize knowledge across tasks\citep{DBLP:conf/emnlp/WangMAKMNADASPK22, DBLP:conf/acl/MishraKBH22, DBLP:conf/emnlp/YeLR21}.

\paragraph{Unified Discriminative Classifier} Previous research, such as the work presented in \citep{DBLP:conf/acl/XuLZZY23}, also attempts to transform classification problems into binary tasks. While this system represents a discriminative classifier approach similar to ours, there are several significant differences. The most notable distinction is that it does not cover zero-shot learning scenarios, whereas our \taskname encompasses the entire range of label occurrences. Additionally, without any supervision, their approach cannot be adapted to diverse tasks, and therefore, is less flexible than ours. Most importantly, this system benchmarks its performance against generative models, rather than comparing it with state-of-the-art (SOTA) systems specifically designed for target classification tasks.

\section{Problem Statement}

Each \taskname~ target task has the following components: 

\textbullet\enspace\textbf{Input $t$}: Versatile text in varied forms, lengths, and domains.

\textbullet\enspace\textbf{Label space $L$}: $L$ contains arbitrary size of labels: $\{\cdots, l_i, \cdots\}$ and an optional \textit{None} label (i.e., all labels in $L$ are incorrect for the input). Within $L$, each label can be either zero-shot, few-shot, or more frequent.

The task of \taskname~is to figure out label $L_s\in L$ that is correct for the input $t$ in the target task, where $|L_s|$ might be zero (i.e., ``\textit{None}'').

\paragraph{Research questions of \taskname:} i) Given that the above formulation encompasses various text classification problems, how can we move away from constructing individual models for each problem, and instead develop a single classifier adept at handling diverse classification settings? ii) Beyond frequently-encountered labels, low-shot labels necessitate additional supervision for effective reasoning. Where can we find such supervision? In the following section, we delve deeper into our approach concerning the universal system and the provided supervisions.

\begin{figure}[t]
\setlength{\belowcaptionskip}{-3pt}
 \setlength{\abovecaptionskip}{-3pt}
 \center
 \includegraphics[width=0.45\textwidth]{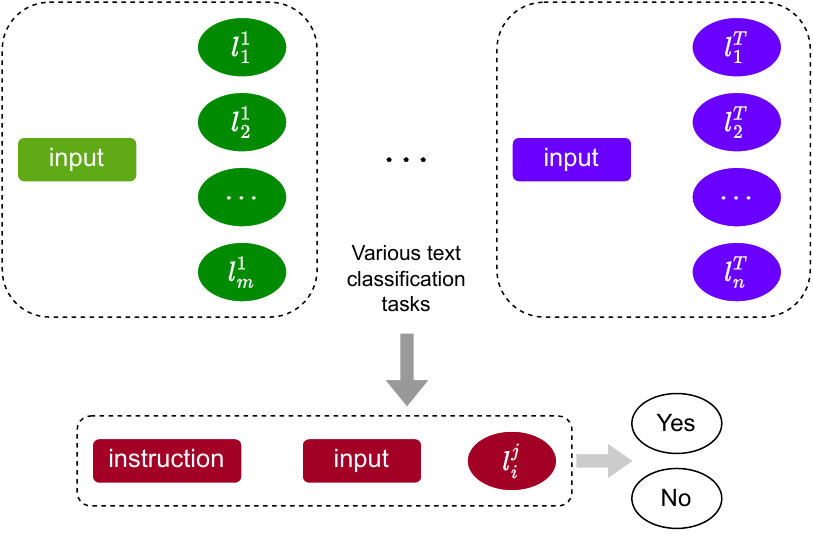}
 \caption{\modelname~unifies $T$ various text classification tasks as an instruction tuning problem. $l_i^j$: the $i$-th label in the $j$-th task. A detailed example is in Appendix \ref{appendix:classification2binary}.}
 \label{fig:model}
\vspace{-3mm}
\end{figure}

\begin{figure*}[t]
\setlength{\belowcaptionskip}{-3pt}
 \setlength{\abovecaptionskip}{-3pt}
 \center
 \includegraphics[width=0.8\textwidth]{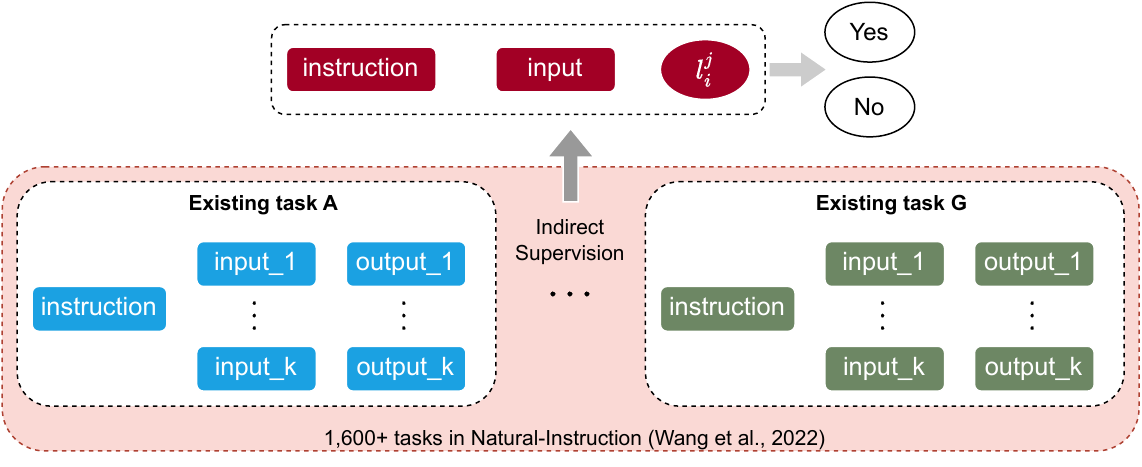}
 \vspace{1mm}
 \caption{\emph{Indirect Supervision} for \modelname. \emph{Indirect Supervision} enables \modelname to learn from diverse tasks in \instrudata before applying this knowledge to a target classification task $j$. A detailed example is in Appendix \ref{appendix:instru2binary} }
 \label{fig:indirect}
         \vspace{-3mm}
\end{figure*}

\section{Methodology}


This section first explains how \modelname~adapts to different classification problems, then introduces the supervision to train it.

\subsection{\modelname~architecture}
We have devised a broad approach that converts any classification task into a unified, instruction-driven binary classification formation. As depicted in Figure \ref{fig:model}, for any text classification task with its set of inputs and labels, we write a short introduction and model it as (\texttt{instruction}, \texttt{input}, \texttt{label}) triplet. The task then becomes determining if the label is appropriate (``Yes'') or not (``No'') given the input under the instruction. This transformation effectively alleviates the frequency gap of the target labels. An example of the conversion can be found in Appendix \ref{appendix:classification2binary}.

\modelname~can support classification tasks with any number of class labels. Instead of mapping labels into numerical representations as traditional supervised classifiers do, we retain the actual label names. 
To pave the way to tackle a variety of low-shot text classification tasks using an instruction-guided approach, two primary challenges arise: i) Ensuring that the model comprehends the instructions, and ii) guiding the model to identify seldom seen or entirely new labels. We will delve deeper into our supervision approaches to address these challenges in the following subsections.

\subsection{Supervision acquisition for low-shot labels}
\taskname relies on \emph{Indirect Supervision} and \emph{Weak Supervision}. We will explain them in this subsection.

\paragraph{Indirect Supervision.} Previous best-performing systems for low-shot text classification have primarily relied on \emph{Indirect Supervision} \textbf{\textit{from a single source task}}. Examples of these source tasks include natural language inference \citep{DBLP:conf/emnlp/YinHR19}, summarization \citep{DBLP:conf/emnlp/LuHZMC22} and passage retrieval \citep{xu-etal-2023-dense}. This approach presents three main drawbacks: i) the usable supervision from the single source task is limited, and there's often a domain mismatch between the source task and the target classification tasks; ii) typically, instances of the target problems need to be reformatted into forms of source tasks to enable zero-shot generalization—a process that's frequently complex; iii) there is not a universally adaptable system to address the \taskname learning, where labels might vary in their occurrences.

In this work, we leverage \emph{Indirect Supervision} from an extensive assortment of NLP tasks. The \instrudata dataset \citep{DBLP:conf/emnlp/WangMAKMNADASPK22} encompasses over 1,600 tasks across 76 categories. Each task is accompanied by instructions and numerous input-output instances (an example of tasks is in Appendix \ref{appendix:instru2binary}). This dataset offers an invaluable source of \emph{Indirect Supervision} for our target \taskname. As in Appendix \ref{appendix:instru2binary}, for every task within the \instrudata dataset, we are presented with the associated instruction as well as the input and the ground truth answer. For each instance selected, we will randomly pick one answer that is different from the ground truth answer within the task, whether the task is generation or classification. As a result, we obtain one positive triplet (\texttt{instruction}, \texttt{input}, \texttt{ground truth}) and one negative triplet (\texttt{instruction}, \texttt{input}, \texttt{random answer}) for each instance in our training dataset as in Figure \ref{fig:indirect}. Our \emph{Indirect Supervision} stems from this dataset training. Such training further significantly mitigates the incongruity exist in varying label frequencies.

When evaluated on target classification tasks, we convert every sample into a triplet-oriented binary instance similarly to the transformation for \instrudata, complemented by a human-written instruction. Given an original instance with text $t$ and positive label $l$, we add an instruction and craft $|L|$ triplets as [(\texttt{instruction}, $t$, $l$), Yes/No] for each label $l$ from the label space $L$, with the gold label as positive and others as negative.

Through this \emph{Indirect Supervision}, minor alterations—be it a word or a few words—can change the class completely. By enabling the model to distinguish the positive and negative classes from marginally changed inputs, we hope the model establishes more distinct decision boundaries.

\paragraph{Weak Supervision for zero-shot labels.} 
In addition to \emph{Indirect Supervision}, we aim to specifically enhance our model's performance on zero-shot labels. Given that we cannot procure annotated instances for these labels, how can we enhance the model's understanding of zero-shot labels without human intervention or labeling? This is where we leverage the capabilities of GPT-3.5 \citep{DBLP:journals/corr/abs-2005-14165} to produce weakly labeled instances. To generate instances for zero-shot labels, we employ in-context learning by randomly selecting demonstrations from few-shot or frequently labeled data. Here's a prompt from the \textit{Maven} event detection dataset, aimed at producing text and event triggers for zero-shot event types: 

\vspace{0mm}
\noindent
\fbox{%
\small
 \begin{minipage}{0.95\linewidth}
   \underline{event type:} \texttt{Competition} \\
   \underline{event trigger:} \texttt{tournament} \\
   \underline{sentence:} \texttt{The final tournament was Played in two stages: the group stage and the knockout stage.} \\
   
   \underline{event type:} \texttt{Motion} \\
   \underline{event trigger:} \texttt{throwing} \\
   \underline{sentence:} \texttt{Simultaneously, Sayhood gained a lock on Rodriguez, throwing him onto the defensive.} \\ 
   
   \underline{event type:} \texttt{Manufacturing}
 \end{minipage}
}
\vspace{0mm}

By exposing GPT-3.5 to event and event statement examples associated with the event type labels ``Competition'' and ``Motion'', we introduce the zero-shot label ``Manufacturing.'' Subsequently, GPT-3.5 generates an event trigger along with an event statement, serving as a weakly supervised instance for this label.

\paragraph{Model selection. } In the main results, we adopt the pre-trained RoBERTa-large model (355M parameters) \citep{DBLP:journals/corr/abs-1907-11692} as our backbone model, given its reliability and high efficiency. However, \modelname can also be extended to different model scales and architectures, such as T5 and GPTs. More experimental details can be found in Section 5.3. 

\paragraph{Training strategy. } 
We first train the backbone model \citep{DBLP:journals/corr/abs-1907-11692} on the transformed binary \instrudata dataset, then fine-tune on the converted triplet instances of downstream \taskname~tasks. The same backbone model will be used in all experiments and baselines.

\section{Experiments}
\subsection{Experimental setting}

\paragraph{Datasets.} 
In this work, we standardize challenging datasets that can cover (i) multiple domains, (ii) various sizes of class labels, and (iii) out-of-domain label scenarios. 
Therefore, we select: \textit{FewRel} \citep{DBLP:conf/emnlp/HanZYWYLS18}, \textit{MAVEN} \citep{DBLP:conf/emnlp/WangWHJHLLLLZ20}, and \textit{RAMS} \citep{DBLP:conf/acl/EbnerXCRD20}, referring to \emph{relation classification}, \emph{event detection}, and \emph{argument role identification} problems respectively. We converted each data set into a format appropriate for \modelname. Few/zero/freq-shot labels are evenly distributed in all three datasets to avoid bias on any group when reporting the overall performance. Details of label distribution can be seen in Table \ref{tab:data}. We rename each resulting dataset as ``[]$_{\small\taskname}$.''


\textbullet\enspace\textbf{FewRel\small$_\taskname$}: \textit{FewRel} is a well-established relation classification dataset where each instance provides a relation statement, two entities from the statement, and their corresponding relation label. Since the test set of \textit{FewRel} is not available, we include 78 relations from its \emph{train} and \emph{dev} and divide them into 26/26/26 as freq/few/zero-shot labels. We randomly select 500/5/0 instances from each freq/few/zero label in the new $train$, and 200 instances from each label in the new $dev$ and $test$.

\textbullet\enspace\textbf{MAVEN\small$_\taskname$}: As an event detection dataset, the event detection task in \textit{MAVEN} includes two steps: detecting the event trigger and predicting the event label from the trigger. In this work, we will focus on the second step, where we assume the event trigger is known and aim to predict the corresponding event label. To make \textit{MAVEN} align with our setting, we reorganize its $train$ and $dev$ sets as follows: since the event label distribution is significantly imbalanced, we select 69 of them who have 400+ instances plus the ``\textit{None}'' label as our label set. Labels are divided into 23/23/23+1 as freq/few/zero-shot labels with ``\textit{None}'' belonging to the zero-shot group. We select 300/5/0 instances from each freq/few/zero label in the new $train$, and 100 instances from each label in the new $dev$ and $test$.



\textbullet\enspace\textbf{RAMS\small$_\taskname$}: 
\textit{RAMS} tackles the task of identifying semantic role labels given the sentence marked with event triggers and argument terms. There are 30 labels that have more than 100 instances; we split them into 10/10/10 for each label group. Similarly, we select 300/5/0 instances from each freq/few/zero label in the new $train$, and 50 instances from each label in the new $dev$ and $test$.

It's noteworthy that while these datasets may not be the largest in scale, they introduce complex NLP challenges that are non-trivial for the latest LLMs. This complexity arises from the need for \emph{advanced reasoning} and dealing with \emph{extensive label spaces}.


\begin{table}[t]
\centering
\setlength{\tabcolsep}{2pt}
 \begin{tabular}{l|lccc}
  & domain & \#freq & \#few  & \#zero \\
 \hline
 FewRel\tiny$_\taskname$ & Wikipedia & 26 & 26 & 26 \\
 MAVEN\tiny$_\taskname$ & Wikipedia & 23 & 23 & 23+1 \\
 RAMS\tiny$_\taskname$ & News articles & 10 & 10 & 10 \\\bottomrule 
 \end{tabular}
 \vspace{-0.5em}
 \caption{Statistics of dataset labels.}
 \label{tab:data}
 \vspace{-1em}
\end{table}

\paragraph{Baselines.} Four typical baselines are included:

\textbullet\enspace\textbf{Multi-way classification (MWC, \citep{DBLP:conf/acl/SoaresFLK19}) }. 
This methodology is the prior SOTA approach for relation classification which designs a special marker for entity terms. We employ this strategy for all three datasets, given that they all contain term features (entity, event trigger, argument, etc.) similar to relation classification.

\textbullet\enspace\textbf{In-context learning with GPT-3.5 (GPT-3.5).}
We create a prompt that includes three demonstrations, two positive and one negative, and each comes with the input, label, and a True/False label that indicates whether the prediction is correct. The specific process can be seen in Appendix \ref{appendix:in-context baseline template}.

\textbullet\enspace\textbf{\emph{Indirect Supervision} from Text Entailment (NLI; \citealt{DBLP:journals/tacl/LiYC22})}. 
NLI is the prior SOTA approach for addressing a zero-shot or few-shot classification with \emph{Indirect Supervision} from merely the NLI  source task. This paradigm uses the input text as the premise and transforms the label into a hypothesis sentence. 

\textbullet\enspace\textbf{Prototypical Prompt learning (PPL; \citealt{DBLP:conf/acl/CuiHDHL22})} 
PPL is the prior SOTA system for few-shot classification. For each of the dataset, we select 500 instances per label during training for prototype learning. For freq and few shot labels, we keep selecting instances from the available instances until we reach the number. For zero-shot labels, we simply put the label itself as the text for the training since we have no instances available.

\paragraph{Implementation details.} We elaborate on our implementation details at different stages here.

\textbullet\enspace\textbf{\emph{Indirect Supervision}}.
Consistent with the original experimental setup and train/test split \citep{DBLP:conf/emnlp/WangMAKMNADASPK22}, we select 100 random instances from each task when compiling the \emph{Indirect Supervision} dataset from Super-NaturalInstruction. Our prefix template follows the previous benchmark strategy, incorporating only the instruction and two positive examples—provided this inclusion doesn't surpass the word limit. When adjusting target classification tasks to fit \modelname, we draft three distinct instruction prompts and present the average outcomes to demonstrate the system's stability. All templates are available in Appendix \ref{appendix:Task Instuctions}.

\textbullet\enspace\textbf{Weak supervision.} We use the ``text-davinci-003'' GPT-3.5 completion model to augment zero-shot instances. For each zero-shot label, we generate 5 instances to serve as \emph{Weak Supervision}.

\textbullet\enspace\textbf{Prediction threshold.} In the NLI baseline and our method, each instance is converted into $|L|$ Yes/No instances, one for each label. We compare the probability of the positive class to assign labels. For \textit{FewRel} and \textit{RAMS}, the label with the highest score is chosen. In \textit{MAVEN}, we introduce a threshold parameter, \emph{t}. If the label receiving the highest probability does not exceed this probability threshold, we assign the label as ``\textit{None}''. We experiment with various values of \emph{t}, ranging from 0.5 to 1, and select the optimal one based on $dev$.

\begin{table*}[t]
   \setlength{\belowcaptionskip}{1pt}
  \setlength{\tabcolsep}{1.85pt}
  \centering
  \begin{tabular}{lccrrrrrrrrrrrr}
    \toprule
    \multirow{2}{*}{Models} & \multicolumn{4}{c}{FewRel\tiny$_\taskname$} & & \multicolumn{4}{c}{RAMS\tiny$_\taskname$} & & \multicolumn{4}{c}{MAVEN\tiny$_\taskname$}\\
    \cline{2-5}
    \cline{7-10}
    \cline{12-15}
     & all & freq & few & zero &
     & all & freq & few & zero &
     & all & freq & few & zero \\ 
    \midrule
    MWC \small\citep{DBLP:conf/acl/SoaresFLK19}
    & 49.82 & 94.23 & 55.23 & 0.0 &
    & 34.47 & \textbf{78.40} & 25.00 & 0.0 &
    & 42.43 & 85.17 & 43.96 & 0.0 \\
    NLI \small\citep{DBLP:journals/tacl/LiYC22}
    & 63.46 & \textbf{95.35} & 48.81 & 46.22 &     
    & 43.07 & 71.40 & 20.40 & 37.40 &
    & 56.31 & \textbf{85.65} & 39.83 & 44.00 \\
    PPL \small\citep{DBLP:conf/acl/CuiHDHL22}
    & 53.23 & 95.15 & \textbf{63.54} &0.0 &
    & 27.13 & 65.00 & 16.20 &0.20 &
    & 46.84 & 85.04 & \textbf{55.52} &0.0 \\
    GPT-3.5
    & 18.24 & 18.22 & 25.33 & 11.17 &
    & 18.19 & 21.21 & 15.15 & 18.19 &
    & 21.43 & 15.15 & 12.12 & 37.50\\
    \midrule
    \modelname
    & \textbf{68.48} & 94.06 & 58.04 & \textbf{53.34} &
    & \textbf{54.70} & 77.00 & \textbf{29.00} & \textbf{58.07} &
    & \textbf{64.96} & 84.32 & 46.64 & \textbf{63.97} \\

    \bottomrule
  \end{tabular}
     \caption{Main results on three benchmark target tasks}
       \label{tab:mian_result}
       \vspace{-2mm}
\end{table*}

\subsection{Results}

    Table \ref{tab:mian_result} compares \modelname~system with baselines. The ``freq'', ``few'', and ``zero'' columns refer to the accuracy of freq-shot, few-shot, and zero-shot labels respectively. Our model consistently outperforms all baselines by a large margin in the ``all'' and ``zero'' dimensions, while occasionally showing slightly lower but on-par performance with the baselines in ``freq'' and ``few''. Analyzing these baselines, we notice that most are ill-suited for the \taskname problem setting, particularly in zero-shot scenarios where annotations are absent. MWC is influenced by the number of label-wise training instances; therefore, its performance, although pretty high for ``freq'', drops quickly to be 0.0 for ``zero''. Similarly, the few-shot prompting (PPL) baseline does well for ``few'' but encounters difficulties with unseen class instances, underscoring the limitations of classification models in the \taskname context. NLI, representing the SOTA in low-shot learning settings, is the only model adept at managing all three types of labels. Nonetheless, when compared with \modelname, NLI's accuracy remains lower in few-shot and zero-shot situations. This indicates that, despite its competency in handling low-shot labels, NLI’s capacity for exploiting limited supervision is inferior to our system.

As one of the most advanced closed-source LLMs, GPT-3.5 shows limited effectiveness in this task, with its performance across three label sets appearing strikingly similar. Although GPT-like models demonstrate robust capabilities in in-context learning, they \emph{fall short in utilizing rich annotations when available} and often \emph{struggle in scenarios with a large label space}. This highlights the flexibility of our \modelname~in handling classification labels of different sizes and occurrences. 

\subsection{Analyses}
In addition to reporting the main results, we further analyze our system in the following dimensions: ($\mathcal{Q}_1$) the individual contribution of \emph{Indirect Supervision} and \emph{Weak Supervision}; ($\mathcal{Q}_2$) is \modelname adaptive to other model scales and architectures? ($\mathcal{Q}_3$) why does ``zero'' show better performance than ``few'' in RAMS$_{\small\taskname}$ and MAVEN$_{\small\taskname}$? ($\mathcal{Q}_4$) Given that our \emph{Indirect Supervision} is derived from a diverse range of NLP tasks in Natural-Instruction \cite{DBLP:conf/emnlp/WangMAKMNADASPK22}, is there a possibility of task leakage? ($\mathcal{Q}_5$) When selecting source tasks for \emph{Indirect Supervision} in instruction-following, which configuration is more effective: having more (diverse) tasks or having more (task-wise) instances? ($\mathcal{Q}_6$) The efficiency of our system. ($\mathcal{Q}_7$) The mistakes our system makes.

\begin{figure}[t]
	\begin{center}
		\centering
		\includegraphics[width=0.45\textwidth]{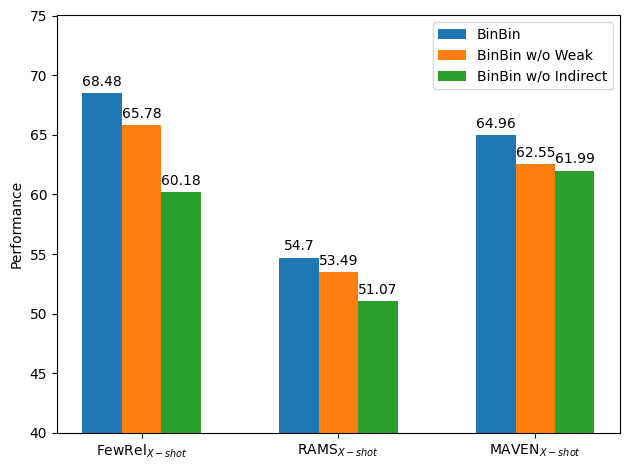}
        \end{center}
  \vspace{-1em}
	\caption{Ablation study of \modelname}
\label{fig:ablation_model}
\vspace{-1em}
\end{figure}

\paragraph{($\mathcal{Q}_1$) Ablation study.} Figure \ref{fig:ablation_model} depicts the ablation study, where either \emph{Indirect Supervision} of \emph{Weak Supervision} is discarded from our system \modelname. Our findings reveal that both supervision sources fulfill complementary roles in the \taskname task. Encouragingly, while their combined usage yields the best results, each type of supervision, on its own, still surpasses the baselines. Such a result underscores the efficiency of our system.

\begin{figure}[t]
	\begin{center}
		\centering
		\includegraphics[width=0.45\textwidth]{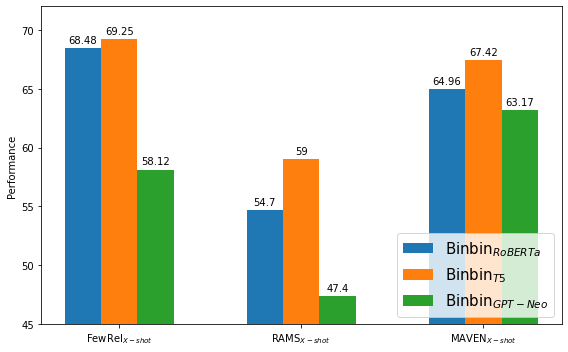}
        \end{center}
  \vspace{-1em}
	\caption{Backbone models across different scales and architectures}
\label{fig:analysis_Q2}
\end{figure}

\paragraph{($\mathcal{Q}_2$) How does \modelname adapt to other model scales and architectures.} Even though we use RoBERTa as our backbone model, \modelname can be adapted to any popular pretrained language model architectures. Besides our main results with RoBERTa-large, an encoder-only transformer with 355M parameters, we also integrate our system into T5-3b \citep{DBLP:journals/jmlr/RaffelSRLNMZLL20} and GPT-Neo 1.3B \citep{gpt-neo}, which are representative models for encoder-decoder and decoder-only transformers, respectively. For RoBERTa, we use the [CLS] token for classification. Similarly, for T5, we only adopt the encoder part and feed the first token into a classification head. For GPT-Neo, since it is a decoder-only model designed for generation tasks, we adopt the last token and add a classification head on top. The results are in Figure \ref{fig:analysis_Q2}. Given the larger parameter size, it is not surprising to see T5-3B outperform RoBERTa across all three datasets. However, GPT-Neo 1.3B consistently underperforms compared to RoBERTa, despite having a similar large parameter size. Considering that both RoBERTa and T5 provide encoder token representations for classification heads, we conclude that decoder-only architectures, such as GPT-Neo, are not as effective in sequence classification.

\paragraph{($\mathcal{Q}_3$) Why do zero-shot labels outperform few-shot labels in the \textit{MAVEN\tiny$_\taskname$} and \textit{RAMS\tiny$_\taskname$} benchmarks? } We observe that this phenomenon applies not only to our system, but also to baselines ``NLI'' and ``GPT-3.5''. We suspect two reasons: i) Some zero-shot labels in \textit{RAMS\tiny$_\taskname$} seem easier upon visual inspection; ii) In \textit{MAVEN\tiny$_\taskname$}, ``None'' is treated as a zero-shot label in the test set, contributing notably due to threshold tuning.





\begin{table}[t]
 \vspace{-0.5em}

\begin{tabular}{l|cccc}
\textbf{}     & all  & freq & few & zero \\ \hline
FewRel\tiny$_\taskname$    & 63.34
     &89.04      & \textbf{60.95}     & 40.04     \\
RAMS\tiny$_\taskname$    &51.64 &\textbf{78.74} &\textbf{30.13} &40.07 \\
MAVEN\tiny$_\taskname$   &63.83 &\textbf{85.68} &\textbf{47.48} &58.57 \\

\end{tabular}
\vspace{-0.5em}
\caption{Results of training \modelname~after deleting top-10 similar tasks from Natural-Instruction. Bold: enhanced performance compared to the pre-deletion state. }
 \label{tab: no_simialr_task_result}
 \vspace{-1em}
\end{table} 



\paragraph{($\mathcal{Q}_4$) Influence of Task Type Overlap.} Although the Natural-Instruction task repository doesn't directly contain our target datasets, we remove the top 10 tasks closest to each target dataset to assess the impact of similar tasks. The measurement is based on cosine similarity between Sentence-BERT \cite{DBLPReimersG19} embeddings of the task definitions in the Natural-Instruction dataset and each \taskname target dataset's instruction.

From Table \ref{tab: no_simialr_task_result}, we can observe that: i) The main decreases when the top-10 similar tasks are deleted happen to zero-shot labels. Recall that we only provided \emph{Weak Supervision} for them; this phenomenon indicates that pretraining on similar source tasks can help diminish the impact of noise in the weakly supervised data. ii) Despite slight decreases in ``all'', our results still surpass baselines in Table \ref{tab:mian_result}, underscoring the value of diverse training tasks. This is further supported by subsequent analysis.


\paragraph{($\mathcal{Q}_5$) Number of Tasks vs Number of Instances.}

\begin{figure}
	\begin{center}
		\centering
		\includegraphics[width=0.45\textwidth]{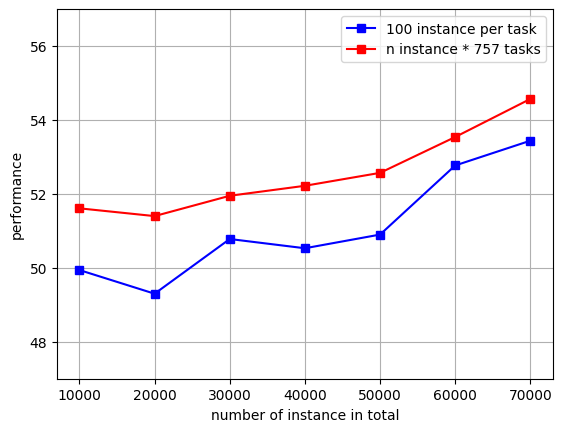}
	\end{center}
  \vspace{-1em}
	\caption{\#instances vs. \#tasks}
\label{fig:ablation_combo}
\vspace{-1em}
\end{figure}


Balancing the number of tasks and the number of instances per task is pivotal in curating instruction-following datasets \cite{DBLP02436}. We wonder, by keeping the total instance count constant, should we have more tasks or more instances per task? We try [100,200,..,700] for the varying number of tasks, each with 100 instances. In total, we have [10,000, 20,000, ... 70,000] instances. Accordingly, for the varying number of instances per task, we have datasets with [10,000/757, 20,000/757, ... 70,000/757] number of instances. The overall instances remain the same in each step. From Figure \ref{fig:ablation_combo}, it's evident that both task count and instance count boost performance. While increasing either is beneficial, having more (diverse) tasks has a greater impact than adding more instances to each task. Given these insights, future work should focus on diversifying the types of tasks exposed to the model, considering data constraints.

\paragraph{($\mathcal{Q}_6$) Efficiency Analysis.}  Efficiency concerns center around the inference stage, where our system converts varied-label classification problems into a binary inference task. This step of \modelname aligns with that of the NLI baseline, the previous SOTA method for low-shot learning. The training in our system takes more time due to pretraining on Natural-Instruction, but during testing, both systems are equally efficient as they make binary decisions for each label. More importantly, using a unified system like BinBin, as opposed to separate systems for different label groups, actually reduces overall training time and computational effort. A more detailed quantitative report in terms of training time and computational resources can be found in Appendix \ref{appendix:efficiency}


\paragraph{($\mathcal{Q}_7$) Error Analysis.} We collect the most typical errors as follows: 

\textbullet\enspace\textbf{Multiple labels make sense}
In datasets with many labels, multiple labels can fit a context, with the model's interpretation sometimes more accurate than the original data. Consider the instance from \textit{RAMS} dataset: ``Many high-ranking figures in companies tied to Skolkovo have also donated to the Clinton Foundation'' While the ground truth label for the argument ``Clinton Foundation'' is ``recipient'', the model strongly suggests ``beneficiary''—a label that is equally justifiable.

\textbullet\enspace\textbf{Bias towards more frequent labels} Models often favor frequently encountered labels in cases of semantic overlap among multiple labels. For example, consider a sentence from the \textit{FewRel} dataset: ``The Spanish - Andorran border runs 64 km between the south of Andorra and northern Spain ( by the autonomous community of Catalonia ) in the Pyrenees Mountains.''. Here, the entities are ``Catalonia'' and ``autonomous community''. Although the gold relation for the two entities is ``instance of'', the model assigns the highest probability to ``part of''—a frequent group label. This suggests that not only does the label share semantic similarities with others, but its frequent occurrence also biases the prediction, especially when many labels lead to potential confusion.

\textbullet\enspace\textbf{identifying reciprocal or inverse relationships} This issue arises when the model struggles to differentiate between roles that represent opposite positions in a given context, such as in a ``receiver'' and ``giver'' scenario while both roles are part of the same transaction, but the model confuses who is who. For instance, in a sentence from \textit{RAMS}:``She was shouting, `I am a terrorist,' and reportedly threatened to blow herself up . ......  he couldn’t believe that the decapitated child ’s head being carried by the woman was real.'' where ``she'' is a ``killer''. However, the model incorrectly labels ``she'' as a ``victim'', demonstrating the difficulty in accurately discerning reciprocal roles.

\section{Conclusion}

This work introduces \taskname, a text classification setting characterized by diverse label occurrences: freq-shot, few-shot, and zero-shot. Our approach, \modelname, leverages \emph{Indirect Supervision} and LLMs' \emph{Weak Supervision} to consistently outperform state-of-the-art methods across three benchmark datasets in various domains.

\section*{Limitation}
The primary limitation of our model is its efficiency, particularly when handling datasets with a large number of labels when converting the original task into a binary task. This results in extended training times and increased computational efforts. It is important to note that this limitation is not an isolated challenge for our model; it aligns with the experiences reported in previous state-of-the-art models. Future work can focus on optimizing the training process to enhance efficiency without compromising the model's performance.


\bibliography{acl}

\begin{thebibliography}{33}
\expandafter\ifx\csname natexlab\endcsname\relax\def\natexlab#1{#1}\fi

\bibitem[{Black et~al.(2021)Black, Gao, Wang, Leahy, and Biderman}]{gpt-neo}
Sid Black, Leo Gao, Phil Wang, Connor Leahy, and Stella Biderman. 2021.
\newblock \href {https://doi.org/10.5281/zenodo.5297715} {{GPT-Neo: Large Scale Autoregressive Language Modeling with Mesh-Tensorflow}}.

\bibitem[{Brown et~al.(2020)Brown, Mann, Ryder, Subbiah, Kaplan, Dhariwal, Neelakantan, Shyam, Sastry, Askell, Agarwal, Herbert{-}Voss, Krueger, Henighan, Child, Ramesh, Ziegler, Wu, Winter, Hesse, Chen, Sigler, Litwin, Gray, Chess, Clark, Berner, McCandlish, Radford, Sutskever, and Amodei}]{DBLP:journals/corr/abs-2005-14165}
Tom~B. Brown, Benjamin Mann, Nick Ryder, Melanie Subbiah, Jared Kaplan, Prafulla Dhariwal, Arvind Neelakantan, Pranav Shyam, Girish Sastry, Amanda Askell, Sandhini Agarwal, Ariel Herbert{-}Voss, Gretchen Krueger, Tom Henighan, Rewon Child, Aditya Ramesh, Daniel~M. Ziegler, Jeffrey Wu, Clemens Winter, Christopher Hesse, Mark Chen, Eric Sigler, Mateusz Litwin, Scott Gray, Benjamin Chess, Jack Clark, Christopher Berner, Sam McCandlish, Alec Radford, Ilya Sutskever, and Dario Amodei. 2020.
\newblock \href {http://arxiv.org/abs/2005.14165} {Language models are few-shot learners}.
\newblock \emph{CoRR}, abs/2005.14165.

\bibitem[{Buda et~al.(2018)Buda, Maki, and Mazurowski}]{DBLP:journals/nn/BudaMM18}
Mateusz Buda, Atsuto Maki, and Maciej~A. Mazurowski. 2018.
\newblock \href {https://doi.org/10.1016/J.NEUNET.2018.07.011} {A systematic study of the class imbalance problem in convolutional neural networks}.
\newblock \emph{Neural Networks}, 106:249--259.

\bibitem[{Cao et~al.(2019)Cao, Wei, Gaidon, Ar{\'{e}}chiga, and Ma}]{DBLP:conf/nips/CaoWGAM19}
Kaidi Cao, Colin Wei, Adrien Gaidon, Nikos Ar{\'{e}}chiga, and Tengyu Ma. 2019.
\newblock \href {https://proceedings.neurips.cc/paper/2019/hash/621461af90cadfdaf0e8d4cc25129f91-Abstract.html} {Learning imbalanced datasets with label-distribution-aware margin loss}.
\newblock In \emph{Advances in Neural Information Processing Systems 32: Annual Conference on Neural Information Processing Systems 2019, NeurIPS 2019, December 8-14, 2019, Vancouver, BC, Canada}, pages 1565--1576.

\bibitem[{Cui et~al.(2022)Cui, Hu, Ding, Huang, and Liu}]{DBLP:conf/acl/CuiHDHL22}
Ganqu Cui, Shengding Hu, Ning Ding, Longtao Huang, and Zhiyuan Liu. 2022.
\newblock \href {https://doi.org/10.18653/v1/2022.acl-long.483} {Prototypical verbalizer for prompt-based few-shot tuning}.
\newblock In \emph{Proceedings of the 60th Annual Meeting of the Association for Computational Linguistics (Volume 1: Long Papers), {ACL} 2022, Dublin, Ireland, May 22-27, 2022}, pages 7014--7024. Association for Computational Linguistics.

\bibitem[{Ebner et~al.(2020)Ebner, Xia, Culkin, Rawlins, and Durme}]{DBLP:conf/acl/EbnerXCRD20}
Seth Ebner, Patrick Xia, Ryan Culkin, Kyle Rawlins, and Benjamin~Van Durme. 2020.
\newblock \href {https://doi.org/10.18653/V1/2020.ACL-MAIN.718} {Multi-sentence argument linking}.
\newblock In \emph{Proceedings of the 58th Annual Meeting of the Association for Computational Linguistics, {ACL} 2020, Online, July 5-10, 2020}, pages 8057--8077. Association for Computational Linguistics.

\bibitem[{Han et~al.(2018)Han, Zhu, Yu, Wang, Yao, Liu, and Sun}]{DBLP:conf/emnlp/HanZYWYLS18}
Xu~Han, Hao Zhu, Pengfei Yu, Ziyun Wang, Yuan Yao, Zhiyuan Liu, and Maosong Sun. 2018.
\newblock \href {https://doi.org/10.18653/v1/d18-1514} {Fewrel: {A} large-scale supervised few-shot relation classification dataset with state-of-the-art evaluation}.
\newblock In \emph{Proceedings of the 2018 Conference on Empirical Methods in Natural Language Processing, Brussels, Belgium, October 31 - November 4, 2018}, pages 4803--4809. Association for Computational Linguistics.

\bibitem[{Li et~al.(2022)Li, Yin, and Chen}]{DBLP:journals/tacl/LiYC22}
Bangzheng Li, Wenpeng Yin, and Muhao Chen. 2022.
\newblock \href {https://doi.org/10.1162/tacl\_a\_00479} {Ultra-fine entity typing with indirect supervision from natural language inference}.
\newblock \emph{Trans. Assoc. Comput. Linguistics}, 10:607--622.

\bibitem[{Li et~al.(2020)Li, Sun, Meng, Liang, Wu, and Li}]{DBLP:conf/acl/LiSMLWL20}
Xiaoya Li, Xiaofei Sun, Yuxian Meng, Junjun Liang, Fei Wu, and Jiwei Li. 2020.
\newblock \href {https://doi.org/10.18653/V1/2020.ACL-MAIN.45} {Dice loss for data-imbalanced {NLP} tasks}.
\newblock In \emph{Proceedings of the 58th Annual Meeting of the Association for Computational Linguistics, {ACL} 2020, Online, July 5-10, 2020}, pages 465--476. Association for Computational Linguistics.

\bibitem[{Liu et~al.(2019)Liu, Ott, Goyal, Du, Joshi, Chen, Levy, Lewis, Zettlemoyer, and Stoyanov}]{DBLP:journals/corr/abs-1907-11692}
Yinhan Liu, Myle Ott, Naman Goyal, Jingfei Du, Mandar Joshi, Danqi Chen, Omer Levy, Mike Lewis, Luke Zettlemoyer, and Veselin Stoyanov. 2019.
\newblock \href {http://arxiv.org/abs/1907.11692} {Roberta: {A} robustly optimized {BERT} pretraining approach}.
\newblock \emph{CoRR}, abs/1907.11692.

\bibitem[{Lou et~al.(2023)Lou, Zhang, Xie, Sun, Ahn, Xu, Su, and Yin}]{DBLP02436}
Renze Lou, Kai Zhang, Jian Xie, Yuxuan Sun, Janice Ahn, Hanzi Xu, Yu~Su, and Wenpeng Yin. 2023.
\newblock {MUFFIN:} curating multi-faceted instructions for improving instruction-following.
\newblock \emph{CoRR}, abs/2312.02436.

\bibitem[{Lu et~al.(2022)Lu, Hsu, Zhou, Ma, and Chen}]{DBLP:conf/emnlp/LuHZMC22}
Keming Lu, I{-}Hung Hsu, Wenxuan Zhou, Mingyu~Derek Ma, and Muhao Chen. 2022.
\newblock \href {https://doi.org/10.18653/v1/2022.findings-emnlp.490} {Summarization as indirect supervision for relation extraction}.
\newblock In \emph{Findings of the Association for Computational Linguistics: {EMNLP} 2022, Abu Dhabi, United Arab Emirates, December 7-11, 2022}, pages 6575--6594. Association for Computational Linguistics.

\bibitem[{Mishra et~al.(2022)Mishra, Khashabi, Baral, and Hajishirzi}]{DBLP:conf/acl/MishraKBH22}
Swaroop Mishra, Daniel Khashabi, Chitta Baral, and Hannaneh Hajishirzi. 2022.
\newblock \href {https://doi.org/10.18653/v1/2022.acl-long.244} {Cross-task generalization via natural language crowdsourcing instructions}.
\newblock In \emph{Proceedings of the 60th Annual Meeting of the Association for Computational Linguistics (Volume 1: Long Papers), {ACL} 2022, Dublin, Ireland, May 22-27, 2022}, pages 3470--3487. Association for Computational Linguistics.

\bibitem[{Obamuyide and Vlachos(2018)}]{obamuyide2018zero}
Abiola Obamuyide and Andreas Vlachos. 2018.
\newblock Zero-shot relation classification as textual entailment.
\newblock In \emph{Proceedings of the first workshop on fact extraction and VERification (FEVER)}, pages 72--78.

\bibitem[{Pouyanfar et~al.(2018)Pouyanfar, Chen, and Shyu}]{DBLP:conf/iri/PouyanfarCS18}
Samira Pouyanfar, Shu{-}Ching Chen, and Mei{-}Ling Shyu. 2018.
\newblock \href {https://doi.org/10.1109/IRI.2018.00064} {Deep spatio-temporal representation learning for multi-class imbalanced data classification}.
\newblock In \emph{2018 {IEEE} International Conference on Information Reuse and Integration, {IRI} 2018, Salt Lake City, UT, USA, July 6-9, 2018}, pages 386--393. {IEEE}.

\bibitem[{Raffel et~al.(2020)Raffel, Shazeer, Roberts, Lee, Narang, Matena, Zhou, Li, and Liu}]{DBLP:journals/jmlr/RaffelSRLNMZLL20}
Colin Raffel, Noam Shazeer, Adam Roberts, Katherine Lee, Sharan Narang, Michael Matena, Yanqi Zhou, Wei Li, and Peter~J. Liu. 2020.
\newblock \href {http://jmlr.org/papers/v21/20-074.html} {Exploring the limits of transfer learning with a unified text-to-text transformer}.
\newblock \emph{J. Mach. Learn. Res.}, 21:140:1--140:67.

\bibitem[{Reimers and Gurevych(2019)}]{DBLPReimersG19}
Nils Reimers and Iryna Gurevych. 2019.
\newblock Sentence-bert: Sentence embeddings using siamese bert-networks.
\newblock In \emph{Proceedings of the 2019 Conference on Empirical Methods in Natural Language Processing and the 9th International Joint Conference on Natural Language Processing, {EMNLP-IJCNLP} 2019, Hong Kong, China, November 3-7, 2019}, pages 3980--3990. Association for Computational Linguistics.

\bibitem[{Sainz et~al.(2022)Sainz, Gonzalez{-}Dios, de~Lacalle, Min, and Agirre}]{DBLP:conf/naacl/SainzGLMA22}
Oscar Sainz, Itziar Gonzalez{-}Dios, Oier~Lopez de~Lacalle, Bonan Min, and Eneko Agirre. 2022.
\newblock \href {https://doi.org/10.18653/v1/2022.findings-naacl.187} {Textual entailment for event argument extraction: Zero- and few-shot with multi-source learning}.
\newblock In \emph{Findings of the Association for Computational Linguistics: {NAACL} 2022, Seattle, WA, United States, July 10-15, 2022}, pages 2439--2455. Association for Computational Linguistics.

\bibitem[{Soares et~al.(2019)Soares, FitzGerald, Ling, and Kwiatkowski}]{DBLP:conf/acl/SoaresFLK19}
Livio~Baldini Soares, Nicholas FitzGerald, Jeffrey Ling, and Tom Kwiatkowski. 2019.
\newblock \href {https://doi.org/10.18653/v1/p19-1279} {Matching the blanks: Distributional similarity for relation learning}.
\newblock In \emph{Proceedings of the 57th Conference of the Association for Computational Linguistics, {ACL} 2019, Florence, Italy, July 28- August 2, 2019, Volume 1: Long Papers}, pages 2895--2905. Association for Computational Linguistics.

\bibitem[{Wang et~al.(2020)Wang, Wang, Han, Jiang, Han, Liu, Li, Li, Lin, and Zhou}]{DBLP:conf/emnlp/WangWHJHLLLLZ20}
Xiaozhi Wang, Ziqi Wang, Xu~Han, Wangyi Jiang, Rong Han, Zhiyuan Liu, Juanzi Li, Peng Li, Yankai Lin, and Jie Zhou. 2020.
\newblock \href {https://doi.org/10.18653/V1/2020.EMNLP-MAIN.129} {{MAVEN:} {A} massive general domain event detection dataset}.
\newblock In \emph{Proceedings of the 2020 Conference on Empirical Methods in Natural Language Processing, {EMNLP} 2020, Online, November 16-20, 2020}, pages 1652--1671. Association for Computational Linguistics.

\bibitem[{Wang et~al.(2022)Wang, Mishra, Alipoormolabashi, Kordi, Mirzaei, Naik, Ashok, Dhanasekaran, Arunkumar, Stap, Pathak, Karamanolakis, Lai, Purohit, Mondal, Anderson, Kuznia, Doshi, Pal, Patel, Moradshahi, Parmar, Purohit, Varshney, Kaza, Verma, Puri, Karia, Doshi, Sampat, Mishra, A, Patro, Dixit, and Shen}]{DBLP:conf/emnlp/WangMAKMNADASPK22}
Yizhong Wang, Swaroop Mishra, Pegah Alipoormolabashi, Yeganeh Kordi, Amirreza Mirzaei, Atharva Naik, Arjun Ashok, Arut~Selvan Dhanasekaran, Anjana Arunkumar, David Stap, Eshaan Pathak, Giannis Karamanolakis, Haizhi~Gary Lai, Ishan Purohit, Ishani Mondal, Jacob Anderson, Kirby Kuznia, Krima Doshi, Kuntal~Kumar Pal, Maitreya Patel, Mehrad Moradshahi, Mihir Parmar, Mirali Purohit, Neeraj Varshney, Phani~Rohitha Kaza, Pulkit Verma, Ravsehaj~Singh Puri, Rushang Karia, Savan Doshi, Shailaja~Keyur Sampat, Siddhartha Mishra, Sujan~Reddy A, Sumanta Patro, Tanay Dixit, and Xudong Shen. 2022.
\newblock \href {https://doi.org/10.18653/v1/2022.emnlp-main.340} {Super-naturalinstructions: Generalization via declarative instructions on 1600+ {NLP} tasks}.
\newblock In \emph{Proceedings of the 2022 Conference on Empirical Methods in Natural Language Processing, {EMNLP} 2022, Abu Dhabi, United Arab Emirates, December 7-11, 2022}, pages 5085--5109. Association for Computational Linguistics.

\bibitem[{Xu et~al.(2023{\natexlab{a}})Xu, Lin, Zhou, Zheng, and Yang}]{DBLP:conf/acl/XuLZZY23}
Haike Xu, Zongyu Lin, Jing Zhou, Yanan Zheng, and Zhilin Yang. 2023{\natexlab{a}}.
\newblock \href {https://doi.org/10.18653/V1/2023.ACL-LONG.589} {A universal discriminator for zero-shot generalization}.
\newblock In \emph{Proceedings of the 61st Annual Meeting of the Association for Computational Linguistics (Volume 1: Long Papers), {ACL} 2023, Toronto, Canada, July 9-14, 2023}, pages 10559--10575. Association for Computational Linguistics.

\bibitem[{Xu et~al.(2022)Xu, Vucetic, and Yin}]{DBLP:journals/corr/abs-2210-14299}
Hanzi Xu, Slobodan Vucetic, and Wenpeng Yin. 2022.
\newblock \href {https://doi.org/10.48550/arXiv.2210.14299} {Openstance: Real-world zero-shot stance detection}.
\newblock \emph{CoRR}, abs/2210.14299.

\bibitem[{Xu et~al.(2023{\natexlab{b}})Xu, Wang, Dong, and Chen}]{xu-etal-2023-dense}
Nan Xu, Fei Wang, Mingtao Dong, and Muhao Chen. 2023{\natexlab{b}}.
\newblock \href {https://aclanthology.org/2023.findings-emnlp.1002} {Dense retrieval as indirect supervision for large-space decision making}.
\newblock In \emph{Findings of the Association for Computational Linguistics: EMNLP 2023}, pages 15021--15033, Singapore. Association for Computational Linguistics.

\bibitem[{Xu et~al.(2023{\natexlab{c}})Xu, Xiao, Liu, Lu, Jing, and Yu}]{DBLP:conf/aaai/XuXLLJY23}
Pengyu Xu, Lin Xiao, Bing Liu, Sijin Lu, Liping Jing, and Jian Yu. 2023{\natexlab{c}}.
\newblock \href {https://doi.org/10.1609/AAAI.V37I9.26259} {Label-specific feature augmentation for long-tailed multi-label text classification}.
\newblock In \emph{Thirty-Seventh {AAAI} Conference on Artificial Intelligence, {AAAI} 2023, Thirty-Fifth Conference on Innovative Applications of Artificial Intelligence, {IAAI} 2023, Thirteenth Symposium on Educational Advances in Artificial Intelligence, {EAAI} 2023, Washington, DC, USA, February 7-14, 2023}, pages 10602--10610. {AAAI} Press.

\bibitem[{Ye et~al.(2021)Ye, Lin, and Ren}]{DBLP:conf/emnlp/YeLR21}
Qinyuan Ye, Bill~Yuchen Lin, and Xiang Ren. 2021.
\newblock \href {https://doi.org/10.18653/v1/2021.emnlp-main.572} {Crossfit: {A} few-shot learning challenge for cross-task generalization in {NLP}}.
\newblock In \emph{Proceedings of the 2021 Conference on Empirical Methods in Natural Language Processing, {EMNLP} 2021, Virtual Event / Punta Cana, Dominican Republic, 7-11 November, 2021}, pages 7163--7189. Association for Computational Linguistics.

\bibitem[{Yin et~al.(2023)Yin, Chen, Zhou, Ning, Chang, and Roth}]{DBLP:conf/acm/0001CZNCR23}
Wenpeng Yin, Muhao Chen, Ben Zhou, Qiang Ning, Kai{-}Wei Chang, and Dan Roth. 2023.
\newblock \href {https://doi.org/10.18653/v1/2023.acl-tutorials.5} {Indirectly supervised natural language processing}.
\newblock In \emph{Proceedings of the 61st Annual Meeting of the Association for Computational Linguistics: Tutorial Abstracts, {ACL} 2023, Toronto, Canada, July 9-14, 2023}, pages 32--40. Association for Computational Linguistics.

\bibitem[{Yin et~al.(2019)Yin, Hay, and Roth}]{DBLP:conf/emnlp/YinHR19}
Wenpeng Yin, Jamaal Hay, and Dan Roth. 2019.
\newblock \href {https://doi.org/10.18653/v1/D19-1404} {Benchmarking zero-shot text classification: Datasets, evaluation and entailment approach}.
\newblock In \emph{Proceedings of the 2019 Conference on Empirical Methods in Natural Language Processing and the 9th International Joint Conference on Natural Language Processing, {EMNLP-IJCNLP} 2019, Hong Kong, China, November 3-7, 2019}, pages 3912--3921. Association for Computational Linguistics.

\bibitem[{Yin et~al.(2021)Yin, Radev, and Xiong}]{DBLP:conf/acl/YinRX21}
Wenpeng Yin, Dragomir~R. Radev, and Caiming Xiong. 2021.
\newblock \href {https://doi.org/10.18653/v1/2021.findings-acl.435} {Docnli: {A} large-scale dataset for document-level natural language inference}.
\newblock In \emph{Findings of the Association for Computational Linguistics: {ACL/IJCNLP} 2021, Online Event, August 1-6, 2021}, volume {ACL/IJCNLP} 2021 of \emph{Findings of {ACL}}, pages 4913--4922. Association for Computational Linguistics.

\bibitem[{Zhang et~al.(2022)Zhang, Zhang, Huang, and Yu}]{DBLP:conf/emnlp/ZhangZHY22}
Haoxing Zhang, Xiaofeng Zhang, Haibo Huang, and Lei Yu. 2022.
\newblock \href {https://doi.org/10.18653/v1/2022.emnlp-main.87} {Prompt-based meta-learning for few-shot text classification}.
\newblock In \emph{Proceedings of the 2022 Conference on Empirical Methods in Natural Language Processing, {EMNLP} 2022, Abu Dhabi, United Arab Emirates, December 7-11, 2022}, pages 1342--1357. Association for Computational Linguistics.

\bibitem[{Zhang et~al.(2020)Zhang, Hashimoto, Liu, Wu, Wan, Yu, Socher, and Xiong}]{DBLP:conf/emnlp/ZhangHLWWYSX20}
Jian{-}Guo Zhang, Kazuma Hashimoto, Wenhao Liu, Chien{-}Sheng Wu, Yao Wan, Philip~S. Yu, Richard Socher, and Caiming Xiong. 2020.
\newblock \href {https://doi.org/10.18653/v1/2020.emnlp-main.411} {Discriminative nearest neighbor few-shot intent detection by transferring natural language inference}.
\newblock In \emph{Proceedings of the 2020 Conference on Empirical Methods in Natural Language Processing, {EMNLP} 2020, Online, November 16-20, 2020}, pages 5064--5082. Association for Computational Linguistics.

\bibitem[{Zhang et~al.(2019)Zhang, Lertvittayakumjorn, and Guo}]{DBLP:conf/naacl/ZhangLG19}
Jingqing Zhang, Piyawat Lertvittayakumjorn, and Yike Guo. 2019.
\newblock \href {https://doi.org/10.18653/v1/n19-1108} {Integrating semantic knowledge to tackle zero-shot text classification}.
\newblock In \emph{Proceedings of the 2019 Conference of the North American Chapter of the Association for Computational Linguistics: Human Language Technologies, {NAACL-HLT} 2019, Minneapolis, MN, USA, June 2-7, 2019, Volume 1 (Long and Short Papers)}, pages 1031--1040. Association for Computational Linguistics.

\bibitem[{Zhao et~al.(2021)Zhao, Wallace, Feng, Klein, and Singh}]{DBLP:conf/icml/ZhaoWFK021}
Zihao Zhao, Eric Wallace, Shi Feng, Dan Klein, and Sameer Singh. 2021.
\newblock \href {http://proceedings.mlr.press/v139/zhao21c.html} {Calibrate before use: Improving few-shot performance of language models}.
\newblock In \emph{Proceedings of the 38th International Conference on Machine Learning, {ICML} 2021, 18-24 July 2021, Virtual Event}, volume 139 of \emph{Proceedings of Machine Learning Research}, pages 12697--12706. {PMLR}.

\end{thebibliography}

\begin{figure*}
	\begin{center}
		\centering
		\includegraphics[width=\textwidth]{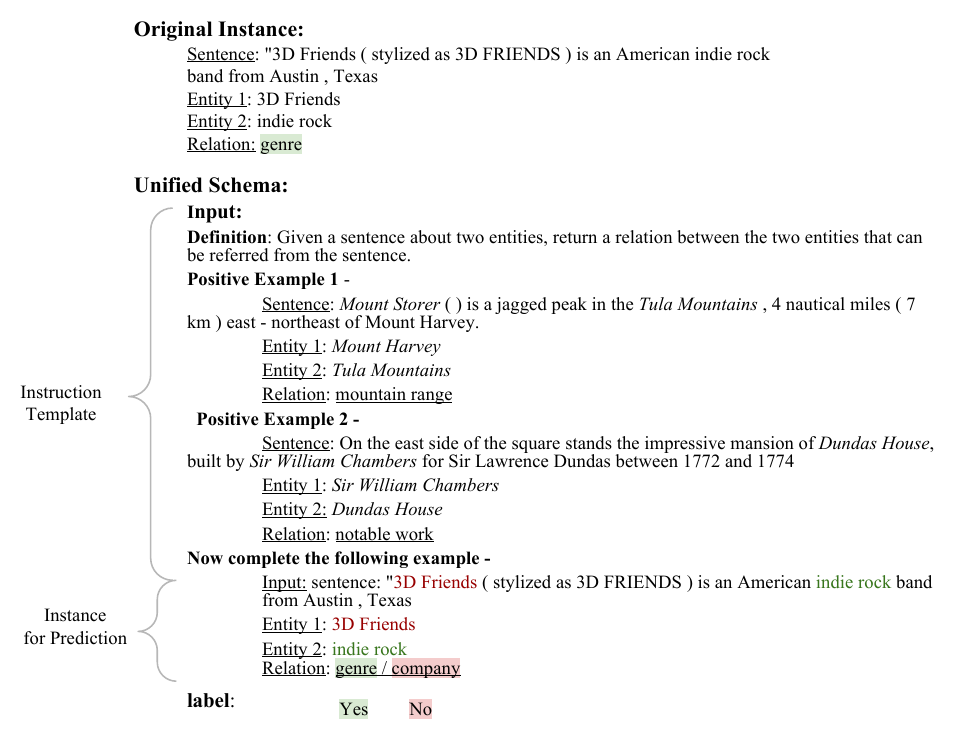}
	\end{center}
  \vspace{-1em}
	\caption{Classification to binary \modelname}
\label{fig:classification2binary}
\vspace{0em}
\end{figure*}

\begin{figure*}
	\begin{center}
		\centering
		\includegraphics[width=0.8\textwidth]{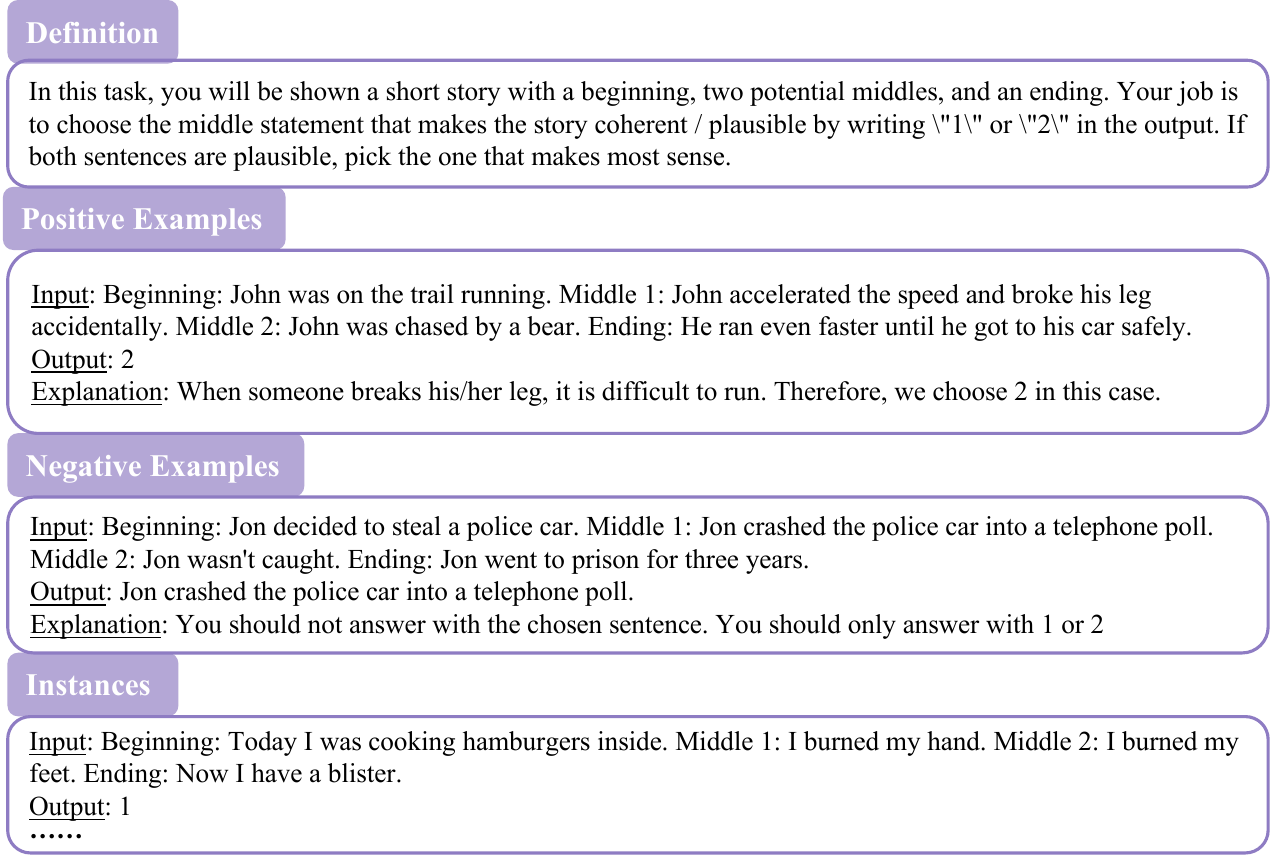}
	\end{center}
  \vspace{-1em}
	\caption{Super-Naturalinstructions task example}
\label{fig:instru_example}
\vspace{0em}
\end{figure*}

\begin{figure*}
	\begin{center}
		\centering
		\includegraphics[width=1\textwidth]{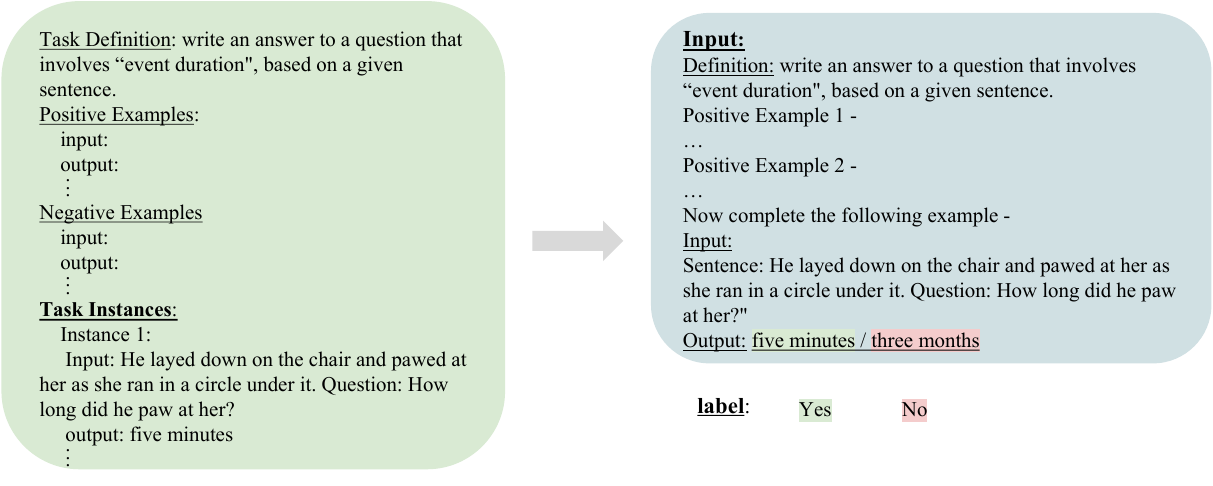}
	\end{center}
  \vspace{-1em}
	\caption{Super-Naturalinstructions to binary \modelname}
\label{fig:instru2binary}
\vspace{0em}
\end{figure*}

\appendix
\section{Appendix}

\subsection{\instrudata to \modelname}
\label{appendix:instru2binary}
We convert \instrudata \citep{DBLP:conf/emnlp/WangMAKMNADASPK22} into our binary schema for the \emph{Indirect Supervision}. \instrudata is a benchmark In-context learning dataset with 757 train tasks and 119 test tasks. Each task includes a definition, several positive/negative demonstrations, and thousands of instances. A task instance from \instrudata is presented in Figure \ref{fig:instru_example}. We select 100 instances from each task and convert them into \modelname schema for \emph{Indirect Supervision} training as shown in Figure \ref{fig:instru2binary}. 

\subsection{\texttt{X-Shot} data to Binbin}
\label{appendix:classification2binary}
As discussed in Section 4.1, each \taskname instance is converted into the unified binary format to align with \modelname. A detailed example from \textit{FewRel} is illustrated in  Figure \ref{fig:classification2binary}.

\subsection{In-context Learning template}
\label{appendix:in-context baseline template}
For the in-context learning baseline, we provide 3 demonstrations, 2 positive ones and 1 negative one, and let GPT-3.5 complete the label of the test instance. A sample template is as follows for \textit{FewRel}: 

\noindent
\fbox{%
 \begin{minipage}{1\linewidth}
   \underline{Sentence:} \texttt{Pan was appointed director of the National Academy (Zhejiang Academy of Fine Arts) by the Kuomintang Ministers} \\
   \underline{Entity 1:} \texttt{Chen Lifu} \\
   \underline{Entity 2:} \texttt{Kuomintang} \\
   \underline{Relation:} \texttt{member of political party} \\
  \underline{Label:} \texttt{Yes} \\
   \\
   \underline{Sentence:} \texttt{Aldo Protti (July 19 ,1920 - August 10 , 1995 ) was an Italian baritone opera singer} \\
   \underline{Entity 1:} \texttt{Aldo Protti} \\
   \underline{Entity 2:} \texttt{baritone} \\
   \underline{Relation:} \texttt{voice type} \\
  \underline{Label:} \texttt{Yes} \\
   \\
   \underline{Sentence:} \texttt{Part of DirectX' Direct3D is used to render three - dimensional graphics in applications } \\
   \underline{Entity 1:} \texttt{DirectX} \\
   \underline{Entity 2:} \texttt{Direct3D} \\
   \underline{Relation:} \texttt{movement} \\
  \underline{Label:} \texttt{No} \\
   \\
   \underline{Sentence:} \texttt{The Suzuki GS500 is an entry level motorcycle manufactured and marketed by the Suzuki Motor Corporation.} \\
   \underline{Entity 1:} \texttt{Suzuki GS500} \\
   \underline{Entity 2:} \texttt{Suzuki Motor Corporation} \\
   \underline{Relation:} \texttt{winner} \\
  \underline{Label:} \\
   \\
 \end{minipage}
  }

\subsection{\modelname Task Instuctions}
\label{appendix:Task Instuctions}
To prove the robustness of our model, we create 3 versions of the task instructions for each of the datasets (\textit{FewRel}, \textit{MAVEN}, \textit{RAMS}) as follows: 

\noindent
\fbox{%
 \begin{minipage}{1\linewidth}
 \textit{FewRel} \\
   \underline{Instruction A:} \texttt{Given a sentence about two entities, return a relation between the two entities that can be inferred from the sentence.} \\
   \underline{Instruction B:} \texttt{Your task is to identify a relationship between two entities mentioned in a given sentence.} \\
   \underline{Instruction C:} \texttt{Identify the relationship between two entities in a given sentence that can be inferred from the sentence.} \\
    \end{minipage}
}

\noindent
\fbox{%
 \begin{minipage}{1\linewidth}
 \textit{RAMS} \\
   \underline{Instruction A:} \texttt{Your task is to identify the role of a specified argument within a given sentence, in relation to an identified event trigger.} \\
   \underline{Instruction B:} \texttt{Identify the role of the argument given the event trigger within the sentence.} \\
   \underline{Instruction C:} \texttt{Identify the role of the argument given the event trigger within the sentence.} \\
    \end{minipage}
}
\noindent
\fbox{%
 \begin{minipage}{1\linewidth}
 \textit{MAVEN} \\
   \underline{Instruction A:} \texttt{Given the sentence and the identified trigger word, determine the most appropriate event category for this trigger.} \\
   \underline{Instruction B:} \texttt{Identify the event type in the sentence associated with the trigger word.} \\
   \underline{Instruction C:} \texttt{Classify the event represented by the trigger word in the context of the following sentence.} \\
   \\

 \end{minipage}
}

\subsection{Efficiency Analysis}
\label{appendix:efficiency} 

\textbullet\enspace\textbf{Time Cost} Our system is trained on NVIDIA A100 GPUs. On a single GPU, it takes 6/30/30 hours on average using the RoBERTa/T5/GPT-Neo model for each task with bf16 precision acceleration. We incorporate packages mainly from Pytorch for the modeling.

\textbullet\enspace\textbf{Memory Cost} The memory requirements for our proposed system include the model parameters and the dataset, similar to other methods and the latest state-of-the-art baseline. The sizes of parameters for the RoBERTa, T5-3B and GPT-Neo models are 355M, 3B, and 1.3B, respectively. For T5, since it is encoder-decoder architecture and we only adopt the encoder, the real memory usage would be 1.5B, half of the original size.

\subsection{ACL ethics code discussion }
\label{appendix:ACL_ethics}
\textbullet\enspace\textbf{Scientific artifacts usage} The existing Scientific artifacts included in this work are RoBERTa, T5 and GPT-Neo model \citep{DBLP:journals/corr/abs-1907-11692, DBLP:journals/jmlr/RaffelSRLNMZLL20,gpt-neo} and 3 NLP classification datasets. The model and datasets used in this work are publicly available for research purposes and do not contain any sensitive information. Our use of existing Scientific artifacts is consistent with their intended usage.

The license, copyright information, the asset we proposed, and terms of use information regarding \modelname, will be specified once the code is released.

\end{document}